\documentclass[letterpaper]{article} 
\usepackage[usenames, dvipsnames]{color}
\usepackage[dvipsnames, svgnames, x11names]{xcolor}

\usepackage[submission]{aaai23}  
\usepackage{times}  
\usepackage{helvet}  
\usepackage{courier}  
\usepackage[hyphens]{url}  
\usepackage{graphicx} 
\usepackage{natbib}  
\usepackage{caption} 
\usepackage{multirow}
\usepackage{booktabs}
\usepackage{subcaption}
\usepackage{amsmath}
\usepackage{bbding}
\usepackage{pifont}
\usepackage{wasysym}
\usepackage{amssymb}
\DeclareMathOperator*{\minimize}{minimize}
\usepackage{algorithmicx,algorithm}
\usepackage[noend]{algpseudocode}
\usepackage{eqparbox}
\renewcommand{\algorithmiccomment}[1]{\bgroup\hfill\#~#1\egroup}

\usepackage{newfloat}
\usepackage{listings}
\DeclareCaptionStyle{ruled}{labelfont=normalfont,labelsep=colon,strut=off} 
\floatstyle{ruled}
\newfloat{listing}{tb}{lst}{}
\floatname{listing}{Listing}

\title{Saliency Guided Contrastive Learning on Scene Images}
\author {
    Meilin Chen\textsuperscript{\rm 1,3*}
    Yizhou Wang\textsuperscript{\rm 1,3}\footnote{Equal contribution. The work was done during an internship at SenseTime.}
    Shixiang Tang\textsuperscript{\rm 2}\footnote{Corresponding author.}
    Feng Zhu\textsuperscript{\rm 3} \\
    Haiyang Yang\textsuperscript{\rm 3,5} 
    Lei Bai\textsuperscript{\rm 4}
    Rui Zhao\textsuperscript{\rm 3,6}
    Donglian Qi\textsuperscript{\rm 1}
    Wanli Ouyang\textsuperscript{\rm 2,4}
}
\affiliations {
    \textsuperscript{\rm 1} Zhejiang University
    \textsuperscript{\rm 2} The University of Sydney
    \textsuperscript{\rm 3} SenseTime Research \\
    \textsuperscript{\rm 4} Shanghai AI Laboratory
    \textsuperscript{\rm 5} Nanjing University \\
    \textsuperscript{\rm 6} Qing Yuan Research Institute, Shanghai Jiao Tong University, Shanghai, China \\
    \{merlinis, yizhouwang, qidl\}@zju.edu.cn stan3906@uni.sydney.edu.au \\
    \{zhufeng, zhaorui\}@sensetime.com baisanshi@gmail.com \\ hyyang@smail.nju.edu.cn wanli.ouyang@sydney.edu.au
}

\usepackage{bibentry}

\begin{document}

\maketitle

\begin{abstract}
Self-supervised learning holds promise in leveraging large numbers of unlabeled data. However, its success heavily relies on the highly-curated dataset, \emph{e.g.,} ImageNet, which still needs human cleaning. Directly learning representations from less-curated scene images is essential for pushing self-supervised learning to a higher level. Different from curated images which include simple and clear semantic information, scene images are more complex and mosaic because they often include complex scenes and multiple objects. Despite being feasible, recent works largely overlooked discovering the most discriminative regions for contrastive learning to object representations in scene images. In this work, we leverage the saliency map derived from the model's output during learning to highlight these discriminative regions and guide the whole contrastive learning. Specifically, the saliency map first guides the method to crop its discriminative regions as positive pairs and then reweighs the contrastive losses among different crops by its saliency scores. Our method significantly improves the performance of self-supervised learning on scene images by \textbf{+1.1}, \textbf{+4.3}, \textbf{+2.2} Top1 accuracy in ImageNet linear evaluation, Semi-supervised learning with 1\% and 10\% ImageNet labels, respectively. We hope our insights on saliency maps can motivate future research on more general-purpose unsupervised representation learning from scene data.
\end{abstract}

\section{Introduction}

Recent self-supervised learning~(SSL) methods~\cite{chen2020simple,grill2020bootstrap,caron2020unsupervised,chen2021exploring} have achieved promising performance that rivals and even surpasses those supervised counterparts on many downstream tasks, such as detection~\cite{chen2022learning} and segmentation. 
However, most of them heavily depend on the highly-curated dataset such as ImageNet~\cite{deng2009imagenet}, where the semantic information in the images is simple and clear, while the enormous and directly available images in the real world are mostly scene images. Compared with the curated images in ImageNet, the semantic information of objects in scene images, such as COCO~\cite{lin2014microsoft}, is complex and mosaic because they usually have multiple objects and complex backgrounds, which is more challenging for self-supervised learning methods to learn the valuable and explicit semantic knowledge. 

\begin{figure}[t]
    \centering
    \includegraphics[width=0.99\linewidth]{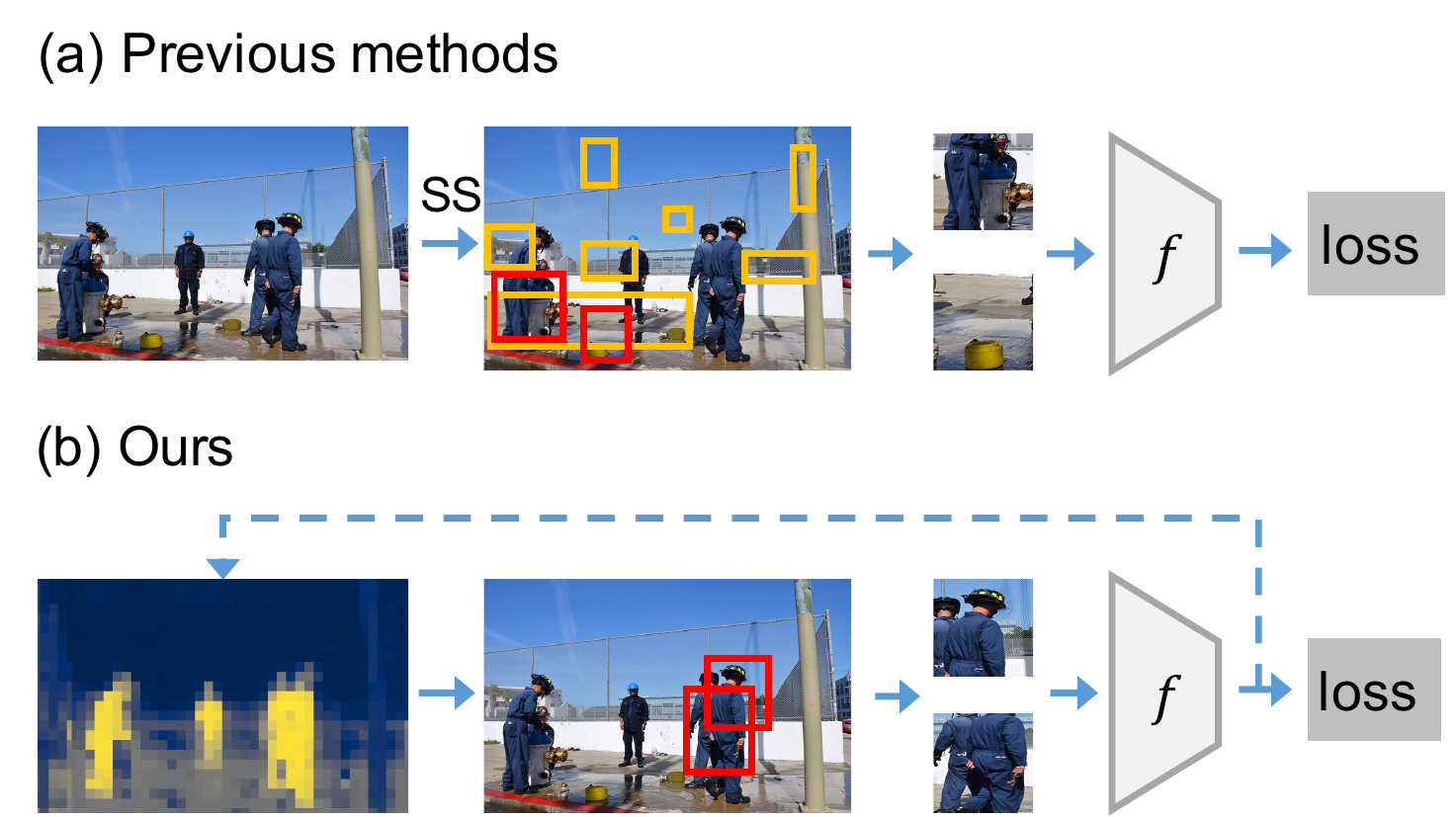}
    \caption{(a) Previous methods use \textbf{Selective Search (SS)} to generate foreground boxes to guide the construction of positive pairs. (b) We propose to use the \textbf{saliency map} instead, which is more dense and accurate, as well as \textbf{processively-refined} with the model learning. Gold and red boxes denote foreground boxes and positive pairs, respectively.}
    \label{fig_teaser}
    \vspace{-1.5em}
\end{figure}

One of the fundamental components of recent self-supervised learning methods on scene images is discovering the discriminative indicators to guide contrastive learning to learn explicit semantic representations of objects. Existing methods~\cite{xie2021unsupervised} leverage Selective Search~\cite{uijlings2013selective} as the indicator method of objectness for contrastive learning by cropping the selected regions to form positive pairs, as shown in Fig.\ref{fig_teaser} (a). Although good performances have been achieved, we argue that such an indicator is noisy and sparse because it is based on a non-deep, heuristic region proposal generation method.

In this paper, we consider the saliency map, which highlights the most important regions for learning object representations, as a more accurate and dense discriminative indicator of guiding contrastive learning on scene images (Fig.\ref{fig_teaser}(b)). First, inspired by~\cite{shi2000normalized, shin2022unsupervised}, the saliency map can be constructed from the self-similarity graph of the learned features of the model. Compared with Selective Search which relies on low-level information of images, saliency maps are more accurate because they can benefit from high-level semantic information in the model outputs and can also be progressively refined along with the model optimization during the pretraining. Second, unlike Search Search only generates bounding boxes, the saliency map is a denser indicator that highlights the discriminative regions in a per-pixel and probabilistic manner. Thus, it can bring denser and more precise information of objectness to guide contrastive learning.

To fully take advantage of the saliency map, we propose a novel \textbf{S}aliency \textbf{G}uided \textbf{C}ontrastive \textbf{L}earning (\textbf{SGCL}) framework for self-supervised learning on the scene images, where saliency generation and model optimization can be processed progressively. The guidance of the saliency map is two folds. First, the saliency map can guide the cropped image generation. Given the saliency map, the discriminative region crops can be generated by selecting the connected regions with high saliency scores. Second, the saliency map can guide the model optimization with a designed saliency-guided loss. Specifically, we weigh the contrastive loss of different crops with their saliency score, which encourages the model to learn object representation from the most discriminative crops containing object information with high saliency scores.
Different from Selective Search generating fixed contrastive crops, the saliency maps in SGCL could be progressively refined to provide better guidance for contrastive learning when the model can learn better discriminative object representations during pretraining.

The contributions of our work are summarized as three-fold:
\textbf{(1)} We propose a saliency-guided contrastive learning framework to learn representations from scene images, where saliency maps are generated via the self-similarity of the learned features, which can be dynamically refined with model pretraining, to guide the construction of positive pairs for contrastive learning in scene images.
\textbf{(2)} We propose a saliency-guided contrastive loss to alleviate the effect of positive pairs with low saliency scores in saliency maps, which can promote representation learning by mitigating the saliency map noise.
\textbf{(3)} Extensive experiments on COCO show that our method significantly improves the performance of self-supervised learning on scene images and pretraining capabilities on several downstream tasks.

\begin{figure*}[t]
    \centering
    \includegraphics[width=0.95\linewidth]{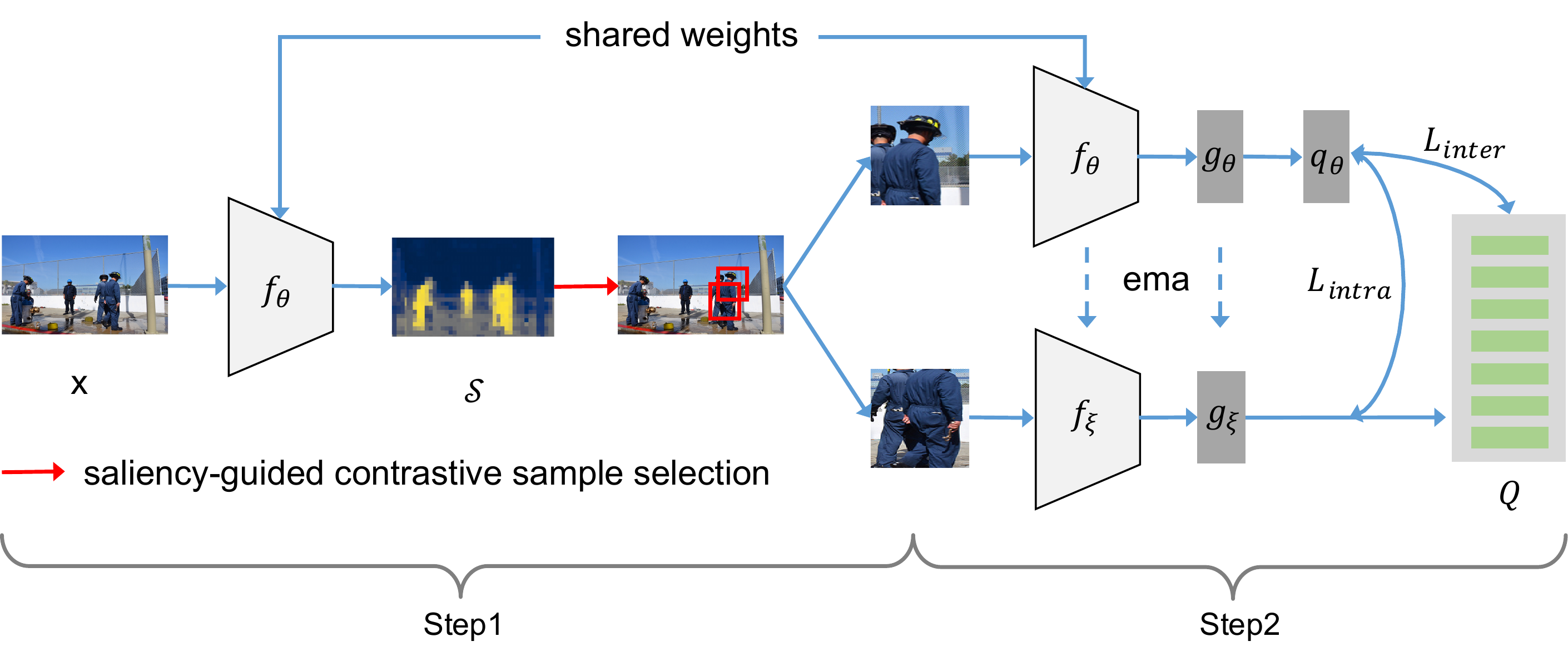}
    \vspace{-1.em}
    \caption{Overview of proposed SGCL. SGCL alternates between two steps. Step1: Dynamically generating saliency map $\mathcal{S}$ to construct positive pairs using encoded features of input data $ \mathbf{x} $, and Step2: optimizing the backbone with saliency-guided contrastive loss $\mathcal{L}_{intra} $ and $ \mathcal{L}_{inter}$. }
    \label{fig_pipeline}
    \vspace{-1.5em}
\end{figure*}

\section{Related Works}
\label{Related Works}
\noindent \textbf{Self-supervised learning on single-centric-object dataset.}
Self-supervised learning aims to obtain object representations from raw data without expensive annotations.
Contrastive learning~\cite{wu2018unsupervised,chen2020improved,he2020momentum,misra2020self,chen2020simple,grill2020bootstrap,caron2020unsupervised,chen2021exploring} has emerged as a promising direction in this field, achieving promising performance that rivals and even surpasses those supervised counterpart. They consider each instance a different class and promote the instance discrimination by forcing representation of different views of the same image closer and spreading representation of views from different images apart. Beyond different views of the same image as the positive sample, NNCLR~\cite{dwibedi2021little} focuses on using nearest neighbors to obtain more diverse positive pairs.

Despite the success in highly curated datasets, these methods heavily rely on the assumption that two random crops for contrastive learning contain explicit and consistent semantic information of objects. When applied to scene scenarios that involve multiple objects of different sizes in complex backgrounds, this assumption no longer holds, which leads to unsatisfactory results. In contrast, SGCL utilizes the saliency map generated by the feature maps to crop views with explicit object information and consistent semantics.

\noindent \textbf{Self-supervised learning on scene images.}
Some previous work \cite{grill2020bootstrap,misra2020self,selvaraju2021casting} naively extend the off-the-shelf contrastive learning methods from single-centric-object datasets to scene datasets, yet they do not acquire satisfactory results because the semantic information in scene images are complex and mosaic. Dense contrastive learning methods~\cite{pinheiro2020unsupervised,liu2020self,wang2021dense,roh2021spatially,henaff2021efficient} leverage local regions of non-iconic images to pretrain the models, yet these methods only work on the specific downstream task, \emph{e.g.,} detection and segmentation. Recently, ORL \cite{xie2021unsupervised} develops a multi-stage framework, where positive pairs are constructed with two cross-image crops generated by Selective Search, showing the potential for pretraining on-scene images. However, its multi-stage design is time and resource-consuming.
Compared with it, our work is a one-stage framework in an end-to-end manner, thus supporting large-scale pretraining.

\noindent \textbf{Construction of positive pairs.}
One of the critical steps in contrastive learning is to construct positive pairs, and most previous works construct positive pairs by random crops of one image. 
Recently, there have been a few works to explore the better construction of positive pairs.  Among them, \citet{mishra2021object, xie2021unsupervised} leverage some heuristic object proposal algorithms to generate foreground boxes to guide the cropping of positive pairs, such as BING~\cite{cheng2014bing} and Selective Search~\cite{uijlings2013selective}. Despite being feasible, the generated boxes are very noisy and are fixed during pretraining. Two closest works~\cite{peng2022crafting, selvaraju2021casting} to this paper use the normalized features and an additional attention loss to learn Grad-CAM attention maps to guide the construction of positive pairs. 

Our work substantially differs from this line of research in three aspects: 
(1) Our work leverage the self-similarity of the learned features to generate saliency maps, which are more accurate and denser to provide semantic information to guide contrastive learning.
(2) Our work directly generate saliency maps using the knowledge learned in contrastive learning. Therefore, our method needs no additional modules or losses, which can be seamlessly integrated with existing contrastive methods.
(3) Beyond guiding the construction of positive pairs, we further propose to guide the model optimization by a designed saliency-guided loss.

\section{Methodology}

To tackle the challenges in pretraining from scene images, we propose a saliency-guided framework as shown in Fig. \ref{fig_pipeline}. The key innovation of the proposed method lies in using the saliency maps to guide the process of pair construction for contrastive learning. The saliency maps are derived from the output of the model and can be progressively refined with the optimization of the backbone network. With this progressive strategy, to avoid training error amplification caused by noisy saliency maps, we further propose a saliency-guided contrastive loss to dynamically weigh the unreliable positive pairs that can be measured by saliency scores.


The training scheme of SGCL alternates between two steps: 
\textbf{(1)} dynamically generating saliency maps to construct the positive pairs using encoded features following the progressive strategy (Sec.~\ref{sec_up}), and \textbf{(2)} optimizing the backbone network with an saliency-guided contrastive loss (Sec.~\ref{sec_un}). 

\subsection{Generating Saliency for Contrastive Learning} 
\label{sec_up}
State-of-the-art unsupervised pretraining methods on scene images leverage random patching~\cite{liu2020self} or Selective Search~\cite{xie2021unsupervised} to guide the construction of contrastive pairs, which in our opinion, is a sub-optimal solution. Instead, we propose to generate saliency maps to construct pairs which are calculated upon high-level feature maps and could encode more semantic information. 
Given a scene images $\mathbf{x}$ and the backbone network $f_\theta$, we extract the feature maps $\mathbb{F}=f_\theta(\mathbf{x})$ to produce the saliency maps $\mathcal{S}$. The positive pairs and their corresponding saliency scores are then generated according to the saliency maps. 

\subsubsection{Generating saliency maps from feature maps}
\label{sec_map}
According to~\cite{shi2000normalized, shin2022unsupervised}, the similarities between foreground and background features are significantly smaller than the similarities between two foreground features or two background features. Therefore, we generate saliency maps by partitioning all spatial features $\mathbf{f}_{i} \in \mathbb{R}^C$ in the encoded feature map $\mathbb{F} \in \mathbb{R}^{H \times W \times C}$ into the foreground set $\mathcal{F}^f$ and the backgrounds set $\mathcal{F}^b$ with the optimization target that the similarities of every two foreground features or the background features are maximized while the similarities between the foreground feature and background feature are minimized. Here, $i \in [1, H\times W]$, $C$ is the number of channels, $H$ and $W$ are the height and width of $\mathbb{F}$. 

Therefore, we formulate the feature partition optimization task as a two-set node partition problem on a self-similarity graph $\mathcal{G}=(\mathcal{V}, \mathcal{E})$, in which the nodes $\mathcal{V}$ denote all spatial features on the feature map $\mathbb{F}$ and the edges $\mathcal{E}$ are the cosine similarity between corresponding features computed by
\begin{equation}
\mathcal{E}_{i, j}= \frac{\left\langle \mathbf{f}_{i}, \mathbf{f}_{j}\right\rangle}{\left\|\mathbf{f}_{i}\right\|_{2} \cdot\left\|\mathbf{f}_{j}\right\|_{2}},
\end{equation}
where $\mathcal{E}_{i, j}$ denotes the edge feature between the spatial feature $\mathbf{f}_{i}$ and $\mathbf{f}_{j}$.

With this favor, the forementioned optimization target can be mathematically formulated as minimizing commonly-adopted Ncut Energy Function~\cite{shi2000normalized} in the graph theory, \emph{i.e.,}  
\begin{equation} 
\footnotesize
\label{eq:Ncut}
    \displaystyle{\minimize_{\mathcal{F}^f,\mathcal{F}^b} \mathbb{E}(\mathcal{F}^f,\mathcal{F}^b) = \minimize_{\mathcal{F}^f,\mathcal{F}^b}\left[ \frac{\operatorname{cut}(\mathcal{F}^f,\mathcal{F}^b)}{\operatorname{cut}(\mathcal{F}^f,\mathcal{V})}+\frac{\operatorname{cut}(\mathcal{F}^f,\mathcal{F}^b)}{\operatorname{cut}(\mathcal{F}^b,\mathcal{V})}\right]},
\end{equation}
where $\operatorname{cut}(A, B)=\sum_{\mathbf{u} \in A, \mathbf{t} \in B} \mathcal{E}_{\mathbf{u}, \mathbf{t}}$ is a function that measures the degree of similarity between two sets $A$ and $B$.
By reducing Eq.~\ref{eq:Ncut} to a continuous optimization problem, the partition probability matrix $\mathbf{Y}\in \mathbb{R}^{H \times W}$ reflecting foreground and background can be calculated by 
\begin{equation}
\footnotesize
\mathbf{Y} = \operatorname{argmin}_{\mathbf{Y}^\top \mathbf{D} \mathbf{1}=0} \frac{\mathbf{Y}^\top(\mathbf{D}-\mathcal{E}) \mathbf{Y}}{\mathbf{Y}^\top \mathbf{D} \mathbf{Y}},
\end{equation}
where $\mathbf{D}$ is the diagonal matrix with $\mathbf{d}_{i}=\sum_{j} \mathcal{E}_{i, j}$ on its diagonal. The detail of the solution process is elaborated in the supplementary material. We leverage the partition probability matrix $\mathbf{Y}$ to define saliency map $\mathcal{S}$ because the higher value in $\mathbf{Y}$ represents the larger probability of the foreground feature, \emph{i.e.,} $\mathcal{S}=\mathbf{Y}$.

\begin{figure}[t]
    \centering
    \includegraphics[width=0.98\linewidth]{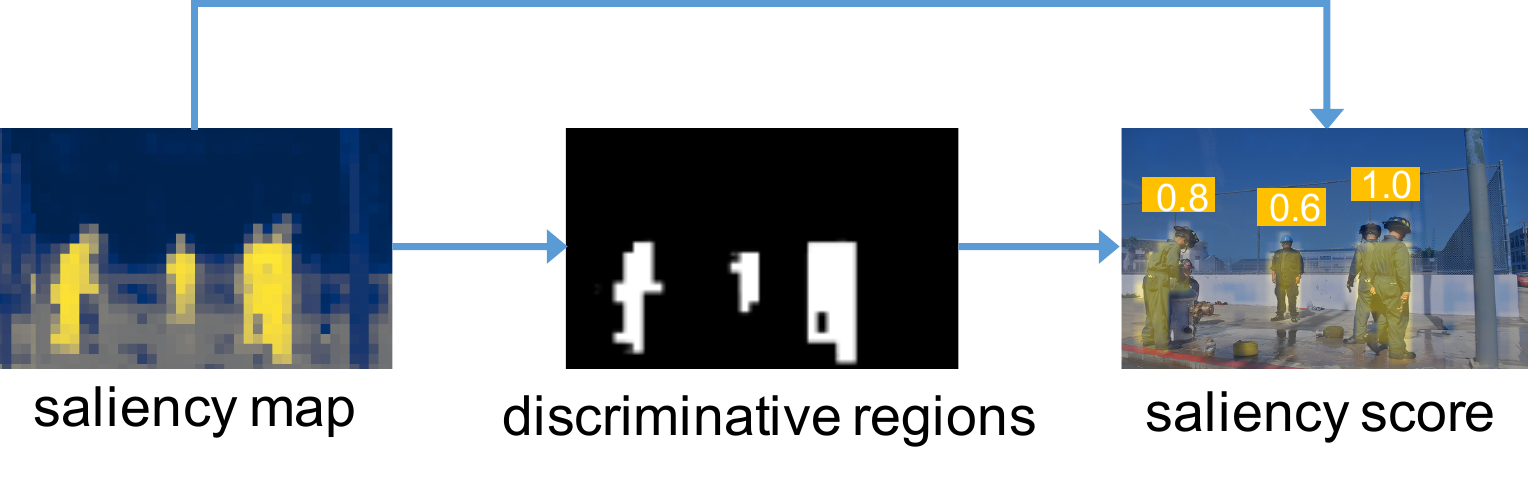}
    \caption{Saliency scores generation from saliency maps. We first partition the discriminative regions out of the saliency maps by thresholding and then compute a score for each region by the relative discriminative strength.}
    \label{fig_proposals_generation}
    \vspace{-1.em}
\end{figure}

\subsubsection{Positive pairs and saliency scores generation from saliency maps.}
\label{sec_scores}

As shown in Figure ~\ref{fig_proposals_generation}, given the saliency maps $\mathcal{S}$ of input data $\mathbf{x}$, we target at generating a set of positive pairs $\{(\mathcal{B}'_1, \mathcal{B}''_1), (\mathcal{B}'_2, \mathcal{B}''_2), ..., (\mathcal{B}'_t, \mathcal{B}''_t)\}$ and their corresponding saliency scores $\{\mathcal{P}_1, \mathcal{P}_2, ..., \mathcal{P}_t\}$.

\noindent \underline{\emph{Discriminative regions generation from saliency maps.}} 
Since the saliency maps highlight the discriminative points, we can partition the discriminative part out of the saliency maps  by thresholding:
\begin{equation} \footnotesize
\mathcal{M}_{ij}= \begin{cases}1, & \text { if } \mathcal{S}_{ij} \textgreater \sigma. \\ 0, & \text { otherwise }\end{cases}
\end{equation}
where $i, j$ are position indice, $\sigma\!=\!\sum_{i=1}^H\sum_{j=1}^W\mathcal{S}_{ij}/HW$ is the average of $\mathcal{S}$, and $\mathcal{M}$ is the bi-partition mask where $\mathcal{M}_{ij}=1$ represents discriminative part. Discriminative regions $\mathcal{B}=\{\mathcal{B}_1, \mathcal{B}_2, ..., \mathcal{B}_K\}$ are generated by grouping each cluster of fully connected points, where $K$ denotes the number of discriminative regions in the bi-partition mask $\mathcal{M}$.

\noindent \underline{\emph{Saliency scores generation from saliency maps.}} 
SGCL can gradually learn better a backbone network to dynamically refine the saliency maps. However, saliency maps inevitably contain noise, and the discriminative regions $\mathcal{B}$ may locate on the backgrounds. To ease this problem, we propose the saliency score for $\mathcal{B}$ using the saliency map $\mathcal{S}$. A higher saliency score usually represents the corresponding region is more reliable. Specifically, we define the saliency score $\mathcal{P}_{i}$ of $\mathcal{B}_{i}$ by the relative discriminative strength:
\begin{equation} \footnotesize
\mathcal{P}_{i} = \frac{\text{max}(\mathcal{S}[\mathcal{B}_{i}])}{\text{max}(\mathcal{S})} ,
\end{equation}
where $\mathcal{S}[\mathcal{B}_{i}]$ denotes the slice of saliency map $ \mathcal{S}$ with the location of $\mathcal{B}_{i}$, $\text{max}(\mathcal{S}[\mathcal{B}_{i}])$ and $\text{max}(\mathcal{S})$ denote the maximum value in the sliced saliency map $\mathcal{S}[\mathcal{B}_{i}]$ and maximum value in the total saliency map $\mathcal{S}$, respecively.

\noindent \underline{\emph{Saliency-guided contrastive sample selection.}}\label{sec:sample_selection}
The saliency score can dynamically weigh each region during the construction of positive pairs, and the regions with higher saliency scores are supposed to be more likely to be chosen to be contrastive samples. Given the regions $\mathcal{B}\!=\!\{\mathcal{B}_1, \mathcal{B}_2, ..., \mathcal{B}_K\}$ and corresponding saliency scores $\mathcal{P}\!=\!\{\mathcal{P}_1, \mathcal{P}_2, ..., \mathcal{P}_K\}$ in a descending order, \emph{i.e.,} $\mathcal{P}_1\ge \mathcal{P}_2 \ge \cdots \ge \mathcal{P}_K$, we choose $t$ regions $\{\mathcal{B}_1, \mathcal{B}_2, ..., \mathcal{B}_t\}$ with replacement for contrastive learning using the weights of saliency scores. Their corresponding saliency scores are $\{\mathcal{P}_1, \mathcal{P}_2, ..., \mathcal{P}_t\}$.

\noindent \underline{\emph{Contrastive pairs construction.}}
Given selected $t$ regions $\{\mathcal{B}_1, \mathcal{B}_2, ..., \mathcal{B}_t\}$ with their saliency scores $\{\mathcal{P}_1, \mathcal{P}_2, ..., \mathcal{P}_t\}$, we aim to construct positive pairs $\{(\mathcal{B}'_1, \mathcal{B}''_1), (\mathcal{B}'_2, \mathcal{B}''_2), ..., (\mathcal{B}'_t, \mathcal{B}''_t)\}$. For each $\mathcal{B}_i$,
\begin{equation} \footnotesize
    (\mathcal{B}'_i, \mathcal{B}''_i) = (\mathcal{T}(\mathcal{B}_i), \mathcal{T}'(\mathcal{B}_i))
\end{equation}
where $\mathcal{T}$ and $\mathcal{T}'$ are the augmentations, including calculating the minimum-area outer rectangle, box jitter and the augmentation methods in BYOL~\cite{grill2020bootstrap}.

\subsection{Saliency-Guided Contrastive Learning} \label{sec_un}
Give an image $\mathbf{x}$, their positive pairs $\{(\mathcal{B}'_1, \mathcal{B}''_1), (\mathcal{B}'_2, \mathcal{B}''_2), ..., (\mathcal{B}'_t, \mathcal{B}''_t)\}$ and corresponding saliency scores $\mathcal{P}=\{\mathcal{P}_1, \mathcal{P}_2, ..., \mathcal{P}_t\}$ generated by Sec.~\ref{sec_up}, we propose to mitigate the saliency map noise by weighing the generated positive pairs according to the sliency scores.


\noindent \textbf{General formulation of saliency-guided contrastive loss.}\label{sec:uncertainty_loss}
To further alleviate the saliency map noise, we propose to leverage the saliency score to give higher weight to more reliable positive pairs. Given $t$ positive pairs $\{(\mathcal{B}'_1, \mathcal{B}''_1), (\mathcal{B}'_2, \mathcal{B}''_2), ..., (\mathcal{B}'_t, \mathcal{B}''_t)\}$ and their corresponding saliency scores $\{\mathcal{P}_1, \mathcal{P}_2, ..., \mathcal{P}_t\}$, the saliency-guided contrastive loss can be formulated as:
\begin{equation} \footnotesize
\begin{aligned}
& \mathcal{L}_{un}\left((\mathcal{B}'_1, \mathcal{B}''_1), (\mathcal{B}'_2, \mathcal{B}''_2), ..., (\mathcal{B}'_t, \mathcal{B}''_t), \mathcal{P}_1, \mathcal{P}_2, ..., \mathcal{P}_t\right) \\
& = \sum_{i=1}^t\mathcal{P}_i^\gamma \mathcal{L}\left(\mathcal{B}'_i, \mathcal{B}''_i\right),
\end{aligned}
\end{equation}
where $\gamma$ is the hyperparameter to determine the degree of weighting and is set as 0.5 in our implementation, and $\mathcal{L}\left(\mathcal{B}'_i, \mathcal{B}''_i\right)$ is the contrastive loss commonly used in BYOL~\cite{grill2020bootstrap}. Similarly, given a positive pair $\mathcal{B}'_i$ and $\mathcal{B}''_i$, the target network with backbone $f_{\xi}$ and projector $g_{\xi}$ provides the regression target to train the online network which consists of backbone $f_{\theta}$, projector $g_{\theta}$ and predictor $q_{\theta}$. Target weights $\xi$ are updated by an exponential moving average of the online parameters $\theta$ with a decay rate $\tau\in [0,1]$. Specifically, the loss function is defined as:
\begin{equation} \footnotesize
\mathcal{L}\left(\mathcal{B}'_i, \mathcal{B}''_i\right) = \frac{\left\langle q_{\theta}(g_{\theta}(f_{\theta}\left(\mathcal{B}'_i\right))), g_{\xi}(f_{\xi}\left(\mathcal{B}''_i\right)\right)\rangle}{\left\|q_\theta(g_\theta(f_{\theta}\left(\mathcal{B}'_i\right)\right))\|_{2} \cdot\left\|g_{\xi}(f_{\xi}\left(\mathcal{B}''_i\right)\right)\|_{2}},
\end{equation}
We name it "intra-" version if $\mathcal{B}'_i$ and $\mathcal{B}''_i$ come from the same image, and name it "inter-" version if $\mathcal{B}'_i$ and $\mathcal{B}''_i$ are generated from different scene images.

\noindent \textbf{Intra-image contrastive loss.} Intra-image contrastive loss pushes two different views close. Given positive pairs $\mathcal{B}'_i$ and  $\mathcal{B}''_i$, mathematically, the intra-image contrastive loss can be formulated as:
\begin{equation} \footnotesize
    \mathcal{L}_{intra} = \sum_{i=1}^t\mathcal{P}_i^\gamma \mathcal{L}\left(\mathcal{B}'_i, \mathcal{B}''_i\right),
\end{equation}

\noindent \textbf{Inter-image contrastive loss.} 
Inspired by NNCLR~\cite{dwibedi2021little}, the inter-image contrastive loss is also used to capture semantic invariance of diverse scales since different viewpoints and object deformations can hardly be created via data augmentations. Specifically, we utilize a memory queue $\mathcal{Q}=\{\mathbf{q}_1, \mathbf{q}_2, ..., \mathbf{q}_r\}$ to store the features extracted in the last few iterations, where $r$ is the size of the queue. We search for $l$-nearest neighbors of $\mathcal{B}'_i$, \emph{i.e.,} $\{\mathcal{N}_1(\mathcal{B}'_i), \mathcal{N}_2(\mathcal{B}'_i), ..., \mathcal{N}_l(\mathcal{B}'_i)\}$, in the memory queue as the positive samples, and the inter-image contrastive loss can be formulated as:
\begin{equation} \footnotesize
    \mathcal{L}_{inter} = \frac{1}{t} \frac{1}{l} \sum_{i=0}^{t-1} \sum_{j=1}^{l} \mathcal{P}_i^\gamma \frac{\left\langle q_{\theta}(g_{\theta}(f_{\theta}\left(\mathcal{B}'_i\right))), \mathcal{N}_j(\mathcal{B}'_i)\right\rangle}{\left\|q_{\theta}(g_{\theta}(f_{\theta}\left(\mathcal{B}'_i\right)\right))\|_{2} \cdot\left\|\mathcal{N}_j(\mathcal{B}'_i)\right\|_{2}},
\end{equation}
where $\mathcal{N}_j(\mathcal{B}_i) \in \mathcal{Q}$ for $i\le t$ and $j \le l$, and $\mathcal{P}_i$ is the corresponding saliency score of $\mathcal{B}_i$. Our inter-image contrastive loss is also guided by the saliency score to mitigate the undesirable optimization by the noisy pairs.

\noindent \textbf{Objective function.} The total loss of proposed SGCL is the combination of saliency-guided intra-image and inter-image contrastive loss with each equal coefficient, \emph{i.e.,}
\begin{equation}
\mathcal{L}_{all} = \mathcal{L}_{intra} + \mathcal{L}_{inter}.
\label{eq_obj_func}
\end{equation}

\noindent\textbf{Computation cost analysis.}
We contribute an efficient technical solution to speed up saliency generation. First, we have efficiently implemented the batch eigen root solution on GPUs with the cupy library. Besides, we parallelize the saliency generation process across all available GPUs, i.e. each GPU generates part of the whole dataset, and then gathers them together. Thus, the computational budget of the saliency map generation process is greatly eased. Quantitatively, with 16 1080 Ti GPUs on COCO (~118k images), the whole training process takes about 70h. Each round of the saliency map generation process takes ~0.5h and we update the saliency map every 100 epochs in the whole 800 epoch schedule. In this case, the saliency map generation process takes only about \textbf{5\%} of the training cost.

\begin{table}[t] \footnotesize
\centering
\resizebox{0.99\linewidth}{!}{
\begin{tabular}{lccccc}
\toprule
Method & Pretrain & ImageNet (Top-1) & VOC07 (mAP)  \\ \midrule
Random & -   & 13.7    & 9.6   \\
Supervised & ImageNet  &  75.9   & 87.5  \\ \midrule
SimCLR & COCO & 50.9  & 78.1  \\
MoCo v2 & COCO & 55.1  & 82.2   \\
NNCLR & COCO & 57.0     & 83.1   \\
BYOL & COCO & 57.8     & 84.5   \\
ORL & COCO & 59.0   & 86.7   \\ 
SGCL & COCO & \textbf{ 59.9 (+0.9)}  &\textbf{86.9 (+0.2)} \\ \midrule
BYOL & COCO+ & 59.6 & 87.0 \\
ORL & COCO+ & 60.7   & 88.6   \\ 
SGCL & COCO+ & \textbf{61.8 (+1.1)}  &\textbf{89.0 (+0.4)} \\ \bottomrule
\end{tabular}
}
\caption{\textbf{Linear evaluation on ImageNet and VOC07.} All SSL methods are trained for 800 epochs  on COCO with ResNet-50 backbone. We report mAP on the VOC07 and top-1 center-crop accuracy on ImageNet.}
\label{tab:linear}
\vspace{-1.em}
\end{table}

\begin{table}[t] \footnotesize
\centering
\resizebox{0.99\linewidth}{!}{
\begin{tabular}{lcccccc}
\toprule
\multirow{2}{*}{Method} & \multirow{2}{*}{Pretrain} & \multicolumn{2}{c}{1\% labels} & \multicolumn{2}{c}{10\% labels} \\
         & & Top-1 & Top-5 & Top-1 & Top-5 \\ \midrule
Random     & - & 1.6      & 5.0      & 21.8      & 44.2      \\
Supervised & ImageNet & 25.4      & 48.4      &  56.4     & 80.4      \\ \midrule
SimCLR & COCO & 23.4      & 46.4      & 52.2      & 77.4      \\
MoCo v2 & COCO & 28.2      & 54.7      & 57.1      & 81.7      \\
NNCLR  & COCO & 28.2      & 54.8      & 57.3     & 82.0      \\
BYOL  & COCO & 28.4      & 55.9      & 58.4      & 82.7      \\
ORL  & COCO & 31.0      & 58.9      & 60.5      & 84.2     \\ 
SGCL &  COCO & \textbf{34.6 (+3.6)}      & \textbf{61.9}      & \textbf{62.2 (+1.7)}      & \textbf{85.0} \\ \midrule
BYOL  & COCO+ & 28.3 & 56.0 & 59.4 & 83.6      \\
ORL  & COCO+ & 31.8 & 60.1 & 60.9 & 84.4     \\
SGCL &  COCO+ & \textbf{36.1 (+4.3)}      & \textbf{63.4}   & \textbf{63.1 (+2.2)}  & \textbf{85.4} \\ \bottomrule
\end{tabular}
}
\caption{\textbf{Reselts of Semi-supervised learning on ImageNet.} All methods are pre-tained on COCO for 800 epochs with ResNet-50 backbone. We fine-tune all models with 1\% and 10\% ImageNet labels, and report both top-1 and top-5 center-crop accuracy on the ImageNet validation set.}
\vspace{-1.5em}
\label{tab_semi}
\end{table}

\section{Experiments}
\subsection{Implementation details} \label{imple_details}
\noindent \textbf{Dataset.}
Following previous works~\cite{xie2021unsupervised}, we pretrain our models on COCO~\cite{lin2014microsoft} train2017 set that contains $\sim$118k images without labels. Compared with the heavily curated object-centric ImageNet dataset, COCO contains more natural and diverse scenes in the wild, which is closer to real-world scenarios and suitable for scene image pretraining.  We also perform self-supervised learning on a larger “COCO+” dataset (COCO \texttt{train2017} set plus COCO \texttt{unlabeled2017} set) to verify whether our method can benefit from more unlabeled scene data. 

\noindent \textbf{Image augmentations.} 
For the saliency map generation in step 1, we rescale all images to 640×640. For the saliency-guided contrastive learning in step 2,  we resize each cropped patch to 96×96, and the subsequent augmentations exactly follow the setting as BYOL~\cite{grill2020bootstrap}.

\noindent \textbf{Network architecture.} Following \cite{xie2021unsupervised}, we adopt ResNet-50 as the default backbone. We use the same MLP projector and predictor as in BYOL~\cite{grill2020bootstrap}: a linear layer with output size 4096 followed by batch normalization~\cite{ioffe2015batch}, rectified linear units~\cite{nair2010rectified}, and a final linear layer with output dimension 256. 

\noindent \textbf{Optimization.} 
We refine the saliency maps with an interval of 100 epochs and initialize model training with random crops as BYOL~\cite{grill2020bootstrap} at the beginning of training. We set $t$ = 4 regions by default. We use the SGD optimizer with a weight decay of 0.0001 and a momentum of 0.9. We adopt the cosine learning rate decay schedule with a base learning rate of 1.0 and the batch size is set to 512 by default. Following ~\cite{xie2021unsupervised}, we train our models for 800 epochs with a warm-up period of 4 epochs. The exponential moving average parameter starts from 0.99 and is increased to 1 during pretraining. All experiments are implemented with 16 NVIDIA 1080ti GPUs. Please refer to supplementary materials for more implementation details.

We omit the citations in the following Tables for better views, and please refer to Sec. \ref{Related Works} for their citations.

\subsection{Transfer to downstream tasks}

\noindent\textbf{Linear evaluation.}
To validate the effectiveness of proposed SGCL, following~\cite{misra2020self,xie2021unsupervised}, we evaluate our learned representation by training a linear classifier on top of
the frozen representation extracted from the network for two datasets: VOC07~\cite{everingham2010pascal} and ImageNet~\cite{deng2009imagenet}. We train linear SVMs on the extracted features using LIBLINEAR package~\cite{fan2008liblinear}. We train on \texttt{trainval} split of VOC07 and evaluate mAP on \texttt{test} split. For ImageNet, we train a linear classifier on \texttt{train} split, and report top-1 center-crop accuracy on the \texttt{val} split. 

The evaluation results are presented in Table~\ref{tab:linear}. First, we can observe that previous conventional SSL methods do not perform well when pretrained on scene images, which validates that conventional SSL heavily relies on the assumption that the pretraining data should have the property of semantic consistency. There is an urgent need for the method to learn good representations from scene images. Second, our proposed SGCL substantially outperforms the BYOL baseline and previous state-of-the-art method ORL~\cite{xie2021unsupervised}.  COCO(+) pretrained SGCL surpasse ORL by \textbf{+0.9} and \textbf{+1.1} Top1 accuracy, indicating that our method is able to learn better representation from scene images. In VOC07 classification setting, COCO(+) pretrained SGCL surpasses official ORL by \textbf{ +0.2} mAP and \textbf{ +0.4} mAP . Notably, with $\sim$ 5x less pretrain data, our COCO+ pretrained SGCL still substantially outperforms ImageNet supervised pretraining by \textbf{+1.5} mAP in VOC07 classification.

\noindent\textbf{Semi-supervised learning.} 
Following the protocol of previous works~\cite{xie2021unsupervised}, we perform semi-supervised learning on ImageNet to validate the effectiveness of our method. Specifically, we use 1\% and 10\% labeled data from ImageNet \texttt{train} split following the image name list provided by SwAV~\cite{caron2020unsupervised}. We then add a classifier on the top of our model to fine-tune the entire models on these two training subsets and report both top-1 and top-5 accuracy on the overall \texttt{val} split of ImageNet. 

As shown in Table~\ref{tab_semi}, our COCO pretrained SGCL substantially outperforms ORL ~\cite{xie2021unsupervised} by \textbf{+3.6} and \textbf{+1.7} of Top-1 accuracy with 1\% and 10\% ImageNet labels, respectively. When it comes to COCO+, SGCL achieves large gains of \textbf{+4.3} and \textbf{+2.2} Top-1 accuracy on 1\% and 10\% settings, respectively. Consistent improvements on these datasets show the superiority of our proposed SGCL when the downstream task is entangled with data-poor challenges.

\begin{table}[t] \footnotesize
\centering
\resizebox{1.0\linewidth}{!}{
\begin{tabular}{lccccccccc}
\toprule
\multirow{2}{*}{Method} & \multirow{2}{*}{Pretrain} &
  \multicolumn{3}{c}{COCO detection} &
  \multicolumn{3}{c}{COCO instance seg.} \\
    &  & AP$^{bb}$ & AP$^{bb}_{50}$ & AP$^{bb}_{75}$ & AP$^{mk}$ & AP$^{mk}_{50}$ & AP$^{mk}_{75}$ \\ \midrule
Random   & - & 32.8   & 50.9   & 35.3   & 29.9   & 47.9   & 32.0   \\
Supervised & ImageNet & 39.7   & 59.5   & 43.3   & 35.9   & 56.6   & 38.6   \\ \midrule
SimCLR  & COCO &  37.0   & 56.8   & 40.3   & 33.7   & 53.8   & 36.1   \\
MoCo v2 & COCO & 38.5   & 58.1   & 42.1   & 34.8   & 55.3   & 37.3   \\
Self-EMD & COCO &  39.3   & 60.1   &  42.8  & -   & -  & - \\
DenseCL & COCO & 39.6   & 59.3   & 43.3   & 35.7   & 56.5   & 38.4   \\
NNCLR & COCO & 39.4   & 59.1   & 43.1   & 35.5   & 56.3   & 38.2   \\
BYOL & COCO & 39.5   & 59.3   & 43.2   & 35.6   & 56.5   & 38.2   \\
ORL & COCO & 40.3  & 60.2   &  44.4  &  36.3   &  57.3  & 38.9   \\ 
SGCL & COCO &\textbf{ 40.7}  & \textbf{60.8}   &  \textbf{44.6}  &  \textbf{36.6}   &  \textbf{57.8}  & \textbf{39.2} \\ \midrule
BYOL & COCO+ & 40.0 & 60.1 & 44.0  & 36.2 & 57.1 & 39.0   \\
ORL & COCO+ & 40.6 & 60.8 & 44.5 & 36.7 & 57.9 & 39.3   \\ 
SGCL & COCO+ &\textbf{41.3}  & \textbf{61.2}   &  \textbf{44.7}  &  \textbf{37.1}   &  \textbf{58.3}  & \textbf{39.7} \\
\bottomrule
\end{tabular}
}
\caption{\textbf{Object detection and instance segmentation on COCO.} All SSL methods are trained for 800 epochs on COCO. We use Mask R-CNN R50-FPN (1$\times$ schedule), and report bounding-box AP (AP$^{bb}$) and mask AP (AP$^{mk}$).}
\label{tab:dense}
\vspace{-1.em}
\end{table}

\begin{table}[t] \footnotesize
\centering
\begin{tabular}{lcccc}
\toprule
$\mathcal{L}_{intra}$ & $\mathcal{L}_{inter}$ & Score & VOC07 \\ \midrule
BYOL & & & 76.9 \\ \midrule
\checkmark &  &   &   78.6 \\
 \checkmark &  \checkmark &  &  81.1  \\
  \checkmark & \checkmark  & \checkmark  &  \textbf{83.2} \\ \bottomrule
\end{tabular}
\caption{Effect of proposed components. $\mathcal{L}_{intra}$ and $\mathcal{L}_{inter}$ intra-image and inter-image contrastive losses. ``Score" represents saliency scores in $\mathcal{L}_{intra}$ and $\mathcal{L}_{inter}$.}
\label{tab_ablation_loss}
\vspace{-1.5em}
\end{table}

\begin{table}[t] \footnotesize
\centering
\begin{tabular}{lcccc}
\toprule
\# NN & 0 & 1 & 5 & 10 \\ \midrule
VOC07 & 80.1 & 82.8 & 83.2 & \textbf{83.4} \\ \bottomrule
\end{tabular}
\caption{Effect of the number of nearest neighbors.}
\label{tab_ablation_knn}
\end{table}

\begin{table}[t] \footnotesize
\centering
\begin{tabular}{lccccc}
\toprule
\# Interval & 5 & 10 & 20 & 50  \\ \midrule
VOC07 & \textbf{83.3} &  83.2 &  83.0 & 82.7  \\
 \bottomrule
\end{tabular}
\caption{Effect of saliency map refinement interval.}
\label{tab_ablation_update}
\vspace{-1.5em}
\end{table}

\noindent\textbf{Object detection and segmentation.}
Apart from classification, we also evaluate our learned representations on detection and instance segmentation tasks to verify the effectiveness of SGCL. Following ORL~\cite{xie2021unsupervised}, we take a Mask R-CNN model~\cite{he2017mask} with R50-FPN backbone~\cite{lin2017feature} implemented in Detectron2~\cite{wu2019detectron2} as our basic model. Specifically, follow the setting in~\cite{tian2020makes}, we \emph{fine-tune} all parameters of the model on \texttt{train2017} split of COCO with the same 1$\times$ schedule, and evaluate on COCO \texttt{val2017} split. For a fair comparison, all the methods are pretrained with the same epochs. 
As shown in Table~\ref{tab:dense}, COCO pretrained SGCL achieves \textbf{+0.4} AP and \textbf{+0.3} AP gains compared with ORL on object detection and instance segmentation, respectively. COCO+ pretrained SGCL consistently surpasses ORL by \textbf{+0.7} AP and \textbf{+0.4} AP.

Besides scene images, our proposed method is also effective when extended to the object-centric dataset, such as ImageNet (refer to supplementary material for experiments).

\subsection{Ablation Study}
In this section, we pretrain our models on COCO and transfer to the VOC07 classification benchmark by default. Unless specified, we pretrain SGCL on COCO for 100 epochs in total with a saliency map refinement interval of 10 epochs.

\noindent\textbf{Effect of proposed components.}
To demonstrate the effectiveness of proposed components in SGCL, Table~\ref{tab_ablation_loss} ablates the effect of intra-image contrastive loss $\mathcal{L}_{intra}$, inter-image contrastive loss $\mathcal{L}_{inter}$ in Eq.~\ref{eq_obj_func}. We observe that SGCL with a simple contrastive loss boosts the BYOL baseline by \textbf{+1.7} mAP, indicating that the proposed SGCL is able to construct more discriminative pairs for pretraining with scene images. By adding inter-image contrastive pairs to leverage the cross-image regions, performance on VOC07 is improved by \textbf{+2.5} mAP. The proposed saliency-guided contrastive loss also brings a \textbf{+1.9} mAP performance gain, showing that saliency-guided contrastive loss can effectively mitigate the effect of noise in saliency maps and boost the representation learning processes.

\noindent\textbf{Effect of the number of nearest neighbors.}
We ablate the number of nearest neighbors $l$ in inter-image contrastive loss $\mathcal{L}_{inter}$ in Table ~\ref{tab_ablation_knn}. We observe that when using only one nearest neighbor, inter-image contrastive loss improves the performance by \textbf{+2.7} mAP. More nearest neighbors bring better performance since more nearest neighbors can provide more diverse positive pairs for contrastive learning. Noticing there is a slight gain from 5 to 10 neighbors, we use 5 neighbors by default.

\noindent\textbf{Effect of saliency map refinement interval.}
From Table~\ref{tab_ablation_update}, we can observe that the smaller refinement interval contributes to the performance since it provides more accurate discriminative regions for contrastive learning. However, the gain brought by the smaller interval decreases when the interval has been small. 

\begin{table}[t] \footnotesize
\centering
\begin{tabular}{cccccc}
\toprule
\# Method & Selective Search & SGCL & GT  \\ \midrule
VOC07 & 80.3 & 83.2 & \textbf{84.7} \\
 \bottomrule
\end{tabular}
\caption{Effect of cropped regions generation methods.}
\label{tab_ablation_proposal}
\end{table}

\noindent\textbf{Comparison with Ground Truth and Selective Search.}
In our SGCL, saliency maps are progressively refined with optimization of the backbone network. To examine this, we compare different positive pairs generated by: (1) Selective Search, (2) our saliency-guided cropping (SCGL), and (3) ground truth (GT). As shown in Table ~\ref{tab_ablation_proposal}, we observe that by using more accurate foreground boxes, from Selective Search to ground truth, the performance is improved by \textbf{+4.4} mAP, indicating that more accurate foreground boxes contribute to the contrastive learning. SGCL achieves \textbf{+2.9} mAP gain compared with Selective Search, showing that with progressive refinement, SGCL can provide better discriminative pairs for contrastive learning.

\noindent\textbf{Effect of the number of the selected regions.}
We ablate the number of of selected regions $t$ in Table ~\ref{tab_ablation_num}. We observe that more selected regions bring better performance since more regions can provide more diverse positive pairs for contrastive learning. In this paper, we set $t$ as 4 by default.

\begin{table}[t] \footnotesize
\centering
\begin{tabular}{cccccc}
\toprule
\# Num & 1 & 2 & 4 & 8 \\ \midrule
VOC07 & 78.7 & 80.8 & 83.2 & \textbf{84.4} \\ \bottomrule
\end{tabular}
\caption{Effect of the number of selected regions.}
\label{tab_ablation_num}
\vspace{-1.5em}
\end{table}

\noindent\textbf{Effect of the saliency generation methods.}
We adapt the other three unsupervised saliency detection methods to our SGCL: (1) heuristic saliency detection (RBD) ~\cite{zhu2014saliency}, (2) normalized feature (NF)  ~\cite{peng2022crafting}, and (3) average pooled embedding as the query embedding (QE) ~\cite{dwibedi2021little}. As shown in Table~\ref{tab_ablation_saliency}, the results show that heuristic saliency leads to a performance drop due to its dependence on low-level features. Besides, our method outperforms the other learnable ones (NF and QE), because our saliency is derived from the self-similarity of the learned features via graph modeling, providing more precise information to guide contrastive learning.

\begin{table}[t] \footnotesize
\centering
\begin{tabular}{ccccc}
\toprule
\# Method & RBD & NF & QE & SGCL \\ \midrule
VOC07 & 79.6 & 81.9 & 82.1 & \textbf{83.2} \\ \bottomrule
\end{tabular}
\caption{Effect of the saliency generation methods.}
\label{tab_ablation_saliency}
\end{table}

\noindent\textbf{Effect of the resolution for saliency generation.}
We ablate the the input resolution for saliency generation in Table \ref{tab_ablation_res}, which shows a performance drop when we reduce the input resolution. Since the saliency is dynamically derived from the encoded feature of backbone, it's spatial scale is reduced by a factor of 32 compared with the input resolution. If the input resolution is too small, the encoded feature is spatially too small to distinguish the adjacent objects and small objects in scene images. We use a resolution of 640 by default as a good trade-off between performance and cost budget.

\begin{table}[t] \footnotesize
\centering
\begin{tabular}{ccccc}
\toprule
\# Resolution &	480	& 640 & 800 \\ \midrule
VOC07 & 82.7 & 83.2 & \textbf{83.5} \\ \bottomrule
\end{tabular}
\caption{Effect of the resolution for saliency generation.}
\label{tab_ablation_res}
\vspace{-1.5em}
\end{table}

\noindent\textbf{Extension to other frameworks.}
Beyond BYOL, we incorporate SGCL with another typical benchmark, SimCLR, and SGCL improves SimCLR of 75.3 mAP to 78.5 mAP by +3.2 mAP in VOC07 classification. This indicates that our proposed SGCL is a general plug-and-play component, which can be seamlessly integrated with existing contrastive methods to boost their representation learning on scene images.

\subsection{Visualization}
\begin{figure}[th]
    \centering
    \vspace{-1.em}
    \includegraphics[width=0.98\linewidth]{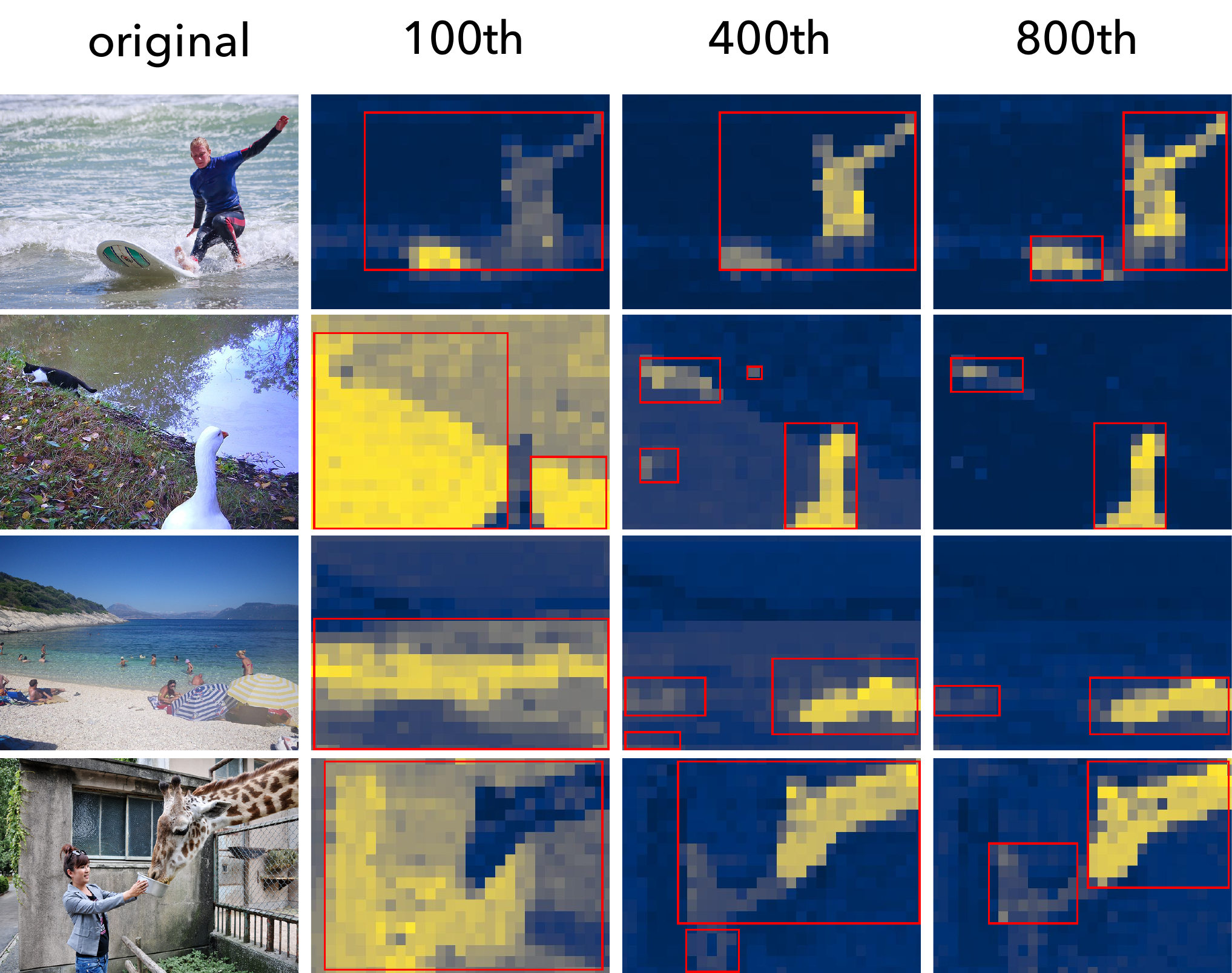}
    \caption{We visualize the saliency maps and crops in different training phases (100-$th$, 400-$th$ and 800-$th$ epoch).}
    \label{fig_vis}
    \vspace{-1.5em}
\end{figure}

We present the qualitative results of the saliency maps and the corresponding crops (Figure ~\ref{fig_vis}). As the training goes on, we can observe that the saliency maps gradually highlight the discriminative regions, leading to more discriminative pairs. This validates that the saliency maps can be progressively refined with the learning of models, which in turn benefits the model learning with more discriminative pairs.

\section{Conclusion}
In this paper, we propose a novel progressive method to learn representations from scene images. Beyond Selective Search, we propose to generate object proposals via the saliency maps, which are derived from encoded semantic features. With this design, the cropped discriminative semantic regions can be progressively refined with the learning of models. With better saliency maps, we can generate better object discriminative region crops for learning object representations by contrastive learning, which in turn benefit the model learning. Besides, we propose a novel saliency-guided contrastive loss to alleviate the effect of crops with low saliency scores. Extensive experiments demonstrate that our method improves the performance of self-supervised learning on scene images. 

\bibliography{aaai23}

\appendix



\section{More Experimental Results}
\noindent\textbf{Pretraining on ImageNet.} 
Beyond scene images, our proposed method is much superior when extended to the object-centric dataset, such as ImageNet.
The results based on 200-epoch pre-training on ImageNet with ResNet-50 are presented in Table \ref{tab_in1k}. Our ImageNet pretrained SGCL outperforms BYOL by \textbf{+3.1} Top1 accuracy.

\noindent\textbf{Comparision between object-centric and scene dataset pretraining.} 
We compare our method pretrained on scene dataset COCO+( $\sim$  240k images, large-scale scene dataset) with self-supervised learning methods pre-trained on ImageNet( $\sim$ 1.28 M images, large-scale object-centric curated dataset). Experimental results show that pretraining on COCO+ is better than pretraining on ImageNet. Specifically, due to the huge data size gap between COCO+ and ImageNet, for fair comparisons, we make sure the number of trained images (epoch x the size of the dataset) are comparable. The results on COCO detection and segmentation with standard 1x schedule are shown in Table \ref{tab_scene_in1k}.

COCO+ pretrained SGCL is better than BYOL trained on ImageNet by \textbf{+0.6} and \textbf{+0.2} mAP in COCO detection and segmentation respectively.
This experimentally validates the merit of the pretraining on scene images as well as the superiority of our SGCL. Besides, pretraining on scene images allows us to learn better features from easily collected off-the-shelf scene data in the wild rather than from the expensive highly curated object-centric images, showing the potential of general and very large-scale vision pretraining like CLIP \cite{radford2021learning}.

\begin{table}[t] 
\centering
\begin{tabular}{lccccc}
\toprule
\# Method & Epoch & Top1 accuracy \\ \midrule
MoCo & 200 & 60.6  \\
SimCLR & 200 & 66.5  \\
MoCov2 & 200 & 67.7  \\
SwAV & 200 & 69.1  \\
SimSiam & 200 & 70.0  \\
BYOL & 200 & 70.5  \\
SGCL & 200 & \textbf{73.6 (+3.1)} \\ \bottomrule
\end{tabular}
\caption{Linear evaluation results on ImageNet. All methods are trained for 200 epochs on ImageNet.}
\label{tab_in1k}
\vspace{-1.5em}
\end{table}

\begin{table*}[tbh] 
\centering
\begin{tabular}{lccccc}
\toprule
Method & Pre-train data & size of dataset & COCO detection(1x) & COCO segmentation(1x) \\ \midrule
MoCo & ImageNet & ~1.28 M & 38.5 & 35.1  \\ 
MoCov2 & ImageNet & ~1.28 M & 38.9 & 35.5 \\
SwAV & ImageNet & ~1.28 M & 38.5 & 35.4 \\
BYOL & ImageNet & ~1.28 M & 40.7 & 36.9 \\
BYOL & COCO+ & ~240k & 40.0 & 36.2 \\
ORL & COCO+ & ~240k & 40.6 & 36.7 \\
SGCL & COCO+ & ~240k & \textbf{41.3 (+0.6)} & \textbf{37.1 (+0.2)} \\ \bottomrule
\end{tabular}
\caption{Comparision between object-centric and scene dataset pretraining. To make sure the number of trained images (epoch x the size of the dataset) are comparable, methods on ImageNet are trained for 200 epochs and methods on COCO+ are trained for 800 epochs.}
\label{tab_scene_in1k}
\end{table*}

\section{More implementation details}

\subsection{Algorithm flow}
The detailed method flow is presented in algorithm~\ref{alg_code}. 
The training scheme of SGCL alternates between two steps: 
\textcolor{red}{\textbf{Step1}}: dynamically generating saliency maps to construct positive pairs using encoded features following the progressive strategy, and \textcolor{blue}{\textbf{Step2}}: optimizing the encoder network with an saliency-guided contrastive loss. 
Note that the notations and references in algorithm~\ref{alg_code} are denoted in the main body of this paper.
We set the threshold of mini-area $s$ as 4 by default.

\begin{algorithm*}[t]
\caption{Saliency Guided Contrastive Learning in Scene Images}
\label{alg_code}
\begin{algorithmic}
\State \textbf{Input:} scene image $I$, update interval $interval$, training epochs $T$, threshold of mini-area $s$
\State initialize $t$ regions $\{\mathcal{B}_1, ..., \mathcal{B}_t\}$  with random cropping \Comment{initialization}
\State initialize $t$ saliency scores $\{\mathcal{P}_1, ..., \mathcal{P}_t\}$  with equal probability 1.0 \Comment{initialization}
\State initialize memory queue $\mathcal{Q}$ with random tensors \Comment{initialization}
\State $epoch = 0$ \Comment{initialization}
\For {$epoch < T$}
    \If {$epoch \neq 0$ and $epoch \% interval = 0$} \Comment{\textcolor{red}{Step1}: generating saliency maps}
        \State $\mathbb{F} = Forward(I)$ \Comment{\textcolor{red}{Step1}: features from last layer of backbone}
        \State $\mathcal{G}=(\mathcal{V}, \mathcal{E}) $ \Comment{\textcolor{red}{Step1}: construct the self-similarity graph}
        \State $ \mathcal{S} \leftarrow (\mathcal{D}-\mathcal{E}) \mathbf{y}=\lambda \mathcal{D} \mathbf{y}$  \Comment{\textcolor{red}{Step1}: the second smallest eigenvector is the saliency maps}
    \EndIf
    \State compute bi-partition mask $\mathcal{M}$ \Comment{\textcolor{red}{Step1}: thresholding following Eq.(4)}
    \State \textcolor{red}{Step1}: delete the discriminative regions whose areas are less than $s$
    \State \textcolor{red}{Step1}: compute saliency scores $\{\mathcal{P}_1, ..., \mathcal{P}_t\}$ following Eq.(5)
    \State \textcolor{red}{Step1}: contrastive sample selection according to saliency scores
    \State \textcolor{red}{Step1}: construct positive pairs $\{(\mathcal{B}'_1, \mathcal{B}''_1),..., (\mathcal{B}'_t, \mathcal{B}''_t)\}$ following Eq.(6)
    \State \textcolor{blue}{Step2}: compute Intra-image and Inter-image loss $\mathcal{L}_{intra}$ and $\mathcal{L}_{inter}$ following Eq.(9-11)
    \State \textcolor{blue}{Step2}: update memory queue $\mathcal{Q}$ 
\EndFor
\end{algorithmic}
\end{algorithm*}

\subsection{Solution process of saliency maps}

As shown by~\cite{shi2000normalized}, the optimization problem of Eq.(2) in the main body of this paper can be equivalently substituted by:

\begin{equation}
\label{eq_Y}
\operatorname{argmin}_{\mathbf{Y}^\top \mathbf{D} \mathbf{1}=0} \frac{\mathbf{Y}^\top(\mathbf{D}-\mathcal{E}) \mathbf{Y}}{\mathbf{Y}^\top \mathbf{D} \mathbf{Y}},
\end{equation}
where $\mathbf{D}$ is the diagonal matrix with $\mathbf{d}_{i}=\sum_{j} \mathcal{E}_{i, j}$ on its diagonal.

Taking $\mathbf{z}=\mathbf{D}^{-\frac{1}{2}}\mathbf{Y}$, S-Eq.~\ref{eq_Y} can be rewrite as:
\begin{equation}
    \mathop{\min}\limits_{\mathbf{z}}{\frac{\mathbf{z}^T\mathbf{D}^{-\frac{1}{2}}(\mathbf{D}-\mathcal{E})\mathbf{D}^{-\frac{1}{2}}\mathbf{z}}{\mathbf{z}^T\mathbf{z}}}.
    \label{eq:energy_z}
\end{equation}

As shown in ~\cite{shi2000normalized}, S-Eq.~\ref{eq:energy_z} is equivalent to the Rayleigh quotient~\cite{golub2013matrix}, which is equivalent to solve $\mathbf{D}^{-\frac{1}{2}}(\mathbf{D}-\mathcal{E})\mathbf{D}^{-\frac{1}{2}}\mathbf{z}=\lambda \mathbf{z}$, where $\mathbf{D}-\mathcal{E}$ is the Laplacian matrix and known to be positive semidefinite~\cite{pothen1990partitioning}.
It can be easily proofed that $\mathbf{z}_0=\mathbf{D}^{-\frac{1}{2}}\mathbf{1}$ is an eigenvector associated to the
smallest eigenvalue of 0, which satisfied the constraint $\mathbf{Y}^T\mathbf{D}\mathbf{1}=0$.  
The second smallest eigenvector, is perpendicular to $\mathbf{z}_0$~\cite{shi2000normalized}. According to the Rayleigh quotient~\cite{golub2013matrix}, $\mathbf{z}_1$, the second smallest eigenvector, is the real valued solution to minimize the energy in S-Eq.~\ref{eq:energy_z},
\begin{equation}
    \mathbf{z}_1 = \mathop{\arg\min}\limits_{\mathbf{z}^T\mathbf{z}_0}{\frac{\mathbf{z}^T\mathbf{D}^{-\frac{1}{2}}(\mathbf{D}-\mathcal{E})\mathbf{D}^{-\frac{1}{2}}\mathbf{z}}{\mathbf{z}^T\mathbf{z}}}.
\end{equation}
Consequently, taking $\mathbf{z}=\mathbf{D}^{-\frac{1}{2}}\mathbf{Y}$, 
\begin{equation}
    \mathbf{Y}_1 = \mathop{\arg\min}\limits_{\mathbf{Y}^T\mathbf{D}\mathbf{1}=0}{\frac{\mathbf{Y}^T(\mathbf{D}-\mathcal{E})\mathbf{Y}}{\mathbf{Y}^T\mathbf{D}\mathbf{Y}}}.
\end{equation}
Therefore, $\mathbf{Y}_1$, the second smallest eigenvector, is the real valued solution that achieves the optimal partition with Ncut energy $\mathbb{E}$. 

We leverage the partition probability matrix,  \emph{i.e.,} the second smallest eigenvector $\mathbf{Y}_1$ as the saliancy map $\mathcal{S}$.

\subsection{positive pairs construction}
Given selected $t$ regions $\{\mathcal{B}_1, \mathcal{B}_2, ..., \mathcal{B}_t\}$ with their saliency scores $\{\mathcal{P}_1, \mathcal{P}_2, ..., \mathcal{P}_t\}$, we aim to construct positive pairs $\{(\mathcal{B}'_1, \mathcal{B}''_1), (\mathcal{B}'_2, \mathcal{B}''_2), ..., (\mathcal{B}'_t, \mathcal{B}''_t)\}$~\footnote{Note that since the instance annotations of the scene images are unavailable, we initialize them with random croppings following BYOL: 8 $\sim $100\% of original image size and 3/4 $\sim $ 4/3 aspect ratio, and initialize each one with equal saliency score 1.0.}. For each $\mathcal{B}_i$ in $\{\mathcal{B}_1, \mathcal{B}_2, ..., \mathcal{B}_t\}$:

\begin{equation}
    (\mathcal{B}'_i, \mathcal{B}''_i) = (\mathcal{T}(\mathcal{B}_i), \mathcal{T}'(\mathcal{B}_i))
\end{equation}
where $\mathcal{T}$ and $\mathcal{T}'$ are the augmentations, including calculating the minimum-area outer rectangle, box jitter and the augmentation methods in BYOL~\cite{grill2020bootstrap}.

To be concrete, for each region $\mathcal{B}_{i}$, we first calculate its minimum-area outer rectangle $\mathbf{b}_{i}$, and then construct positive object-instance pairs $\mathcal{B}_i^{'}$ and $\mathcal{B}_i^{''}$ by cropping around its original location. Specifically, with the box $\mathbf{b}_{i}$, we obtain the cropping coordinates $(\hat{x}, \hat{y}, \hat{h}, \hat{w}) $ with the following principles: \textbf{(1)} a random cropping size $\mathbf{s}$ between the 8 $\sim $100\% of \textbf{box size}, \textbf{(2)} a random cropping center within the box $\mathbf{b}_{i}$, and \textbf{(3)} a random cropping aspect ratio $\mathbf{r}$ between 3/4 and 4/3. Formally, the cropping coordinates of box $\mathbf{b}_{i}$ are generated as follows,

\begin{equation}
(\hat{x}, \hat{y}, \hat{h}, \hat{w})=\mathbb{R}(\mathbf{s}, \mathbf{r}, \mathbf{b}_{i}),
\end{equation}

where $\mathbb{R}$ refers to the rules above. The cropped patch is then augmented to be one view following BYOL~\cite{grill2020bootstrap}.

\section{More Visualization}
\begin{figure*}[tbh]
    \centering
    \includegraphics[width=0.98\linewidth]{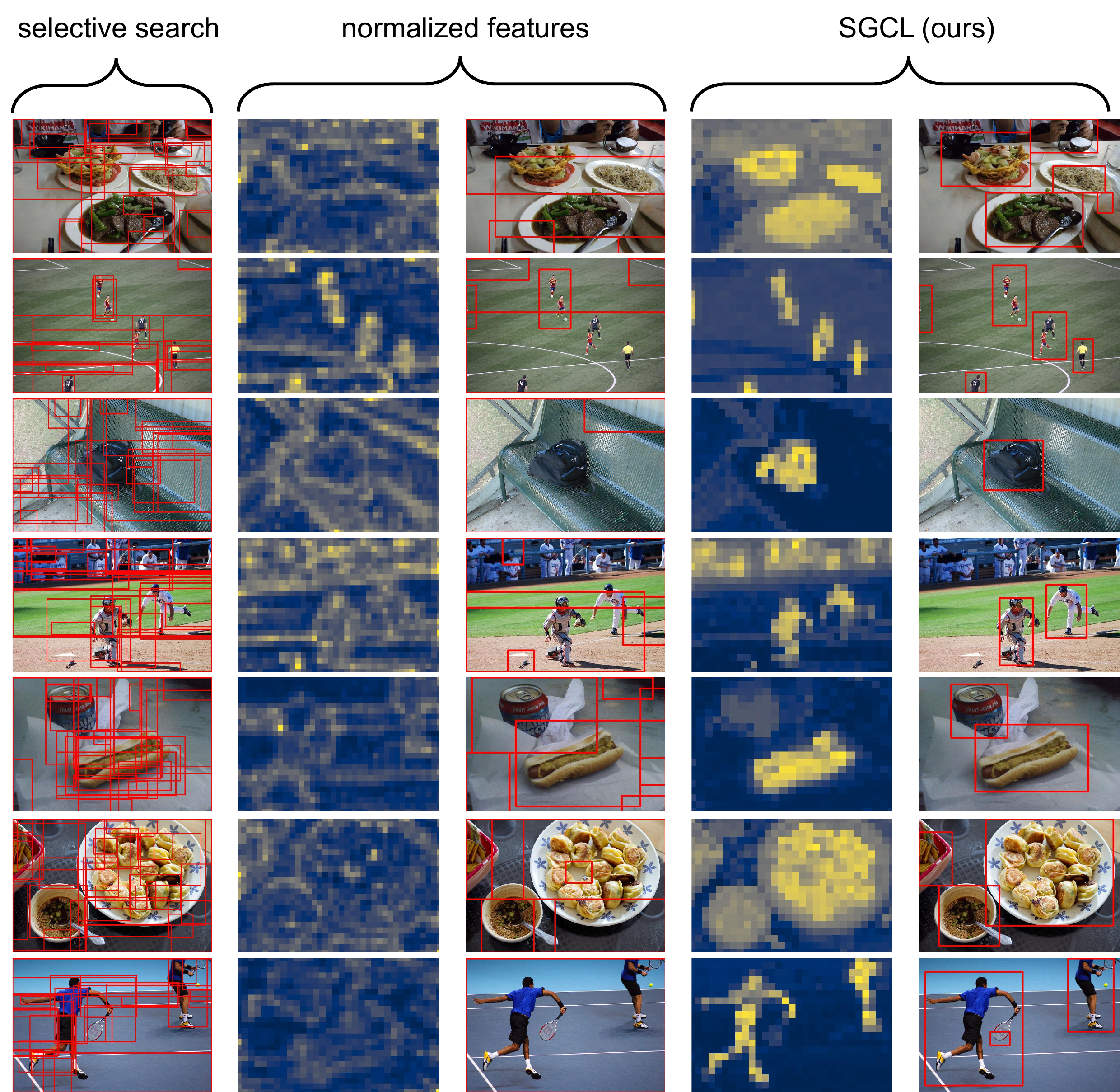}
    \caption{With the same model weights, we visualize more qualitative results: (1) the top 30 croppings generated by the selective search, (2) the heat maps and croppings via normalized features~\cite{peng2022crafting}, and (3) the saliency maps and the corresponding croppings of our proposed SGCL. We can observe that by leveraging the self-similarity of the learned features, our proposed SGCL largely highlights the discriminative objects in the scene images, leading to generating more semantic positive pairs, which in turn benefits the model learning.}
    \label{fig_vis_more}
\end{figure*}

In Figure ~\ref{fig_vis_more}, we present more qualitative results: (1) the top 30 croppings generated by selective search, (2) the heat maps and croppings via normalized features~\cite{peng2022crafting}, and (3) the saliency maps and the corresponding croppings of our proposed SGCL. We can observe that crops generated by selective search and normalized features~\cite{peng2022crafting} are very noisy and most of them are oriented with backgrounds. Compared with them, our proposed SGCL largely highlights the discriminative objects in the scene images, leading to generating more semantic positive pairs, which in turn benefits the model learning.

\end{document}